\documentclass[letterpaper, 10 pt, conference]{ieeeconf}  

\IEEEoverridecommandlockouts                              

\usepackage[utf8]{inputenc}
\usepackage{leftidx}
\usepackage{bm}
\usepackage{amssymb,amsmath,amsfonts,empheq}
\usepackage{graphicx}
\usepackage{algorithm}
\usepackage[noend]{algpseudocode}
\usepackage{type1cm}
\usepackage{pifont}
\usepackage{lettrine}
\usepackage{color}
\usepackage{cite}
\usepackage{breqn}
\usepackage{tikz}
\usepackage{array}
\usepackage{import}

 \usepackage{microtype}

\newtheorem{remrk}{Remark}
\newtheorem{lem}{Lemma}

\usepackage{multirow}

\newcommand{\sclerp}[1]{\text{ScLERP}\left( #1 \right)}

\newcommand{\trans}[1]{\mathfrak T \left( #1 \right)}

\global\long\def\dq#1{\underline{\boldsymbol{#1}}}%

\global\long\def\quat#1{\boldsymbol{#1}}%

\global\long\def\mymatrix#1{\boldsymbol{#1}}%

\global\long\def\myvec#1{\boldsymbol{#1}}%

\global\long\def\dual{\varepsilon}%

\global\long\def\norm#1{\left\Vert #1\right\Vert }%

\global\long\def\hamilton#1#2{\overset{#1}{\operatorname{\mymatrix H}}\left(#2\right)}%

\global\long\def\hami#1{\overset{#1}{\operatorname{\mymatrix H}}}%

\global\long\def\getp#1{\operatorname{\mathcal{P}}\left(#1\right)}%

\global\long\def\getd#1{\operatorname{\mathcal{D}}\left(#1\right)}%

\global\long\def\real#1{\operatorname{\mathrm{Re}}\left(#1\right)}%

\global\long\def\vector{\operatorname{vec}}%

\global\long\def\realset{\ensuremath{\mathbb{R}}}%

\global\long\def\SE#1{\ensuremath{SE(#1)}}%

\global\long\def\unitquatgroup{\ensuremath{\text{Spin}(3)}}%

\global\long\def\unitdualquatgroup{\text{Spin}(3)\ltimes\mathbb{R}^{3}}%

\title{\textbf{Coordinate Invariant User-Guided Constrained Path Planning with Reactive Rapidly Expanding Plane-Oriented
Escaping Trees}}

\author{Riddhiman Laha$^{*,1}$, Ruiai Sun$^{*,1}$, Wenxi Wu$^{*,1}$, Dasharadhan Mahalingam$^{2}$, \\  Nilanjan Chakraborty$^{2}$, Luis F.C. Figueredo$^{*,1}$, and Sami Haddadin$^{1}$\thanks{
$^\ast$All authors contributed equally}%
\thanks{$^{1}$The authors are with Munich Institute of Robotics \& Machine Intelligence, Technische Universität München (TUM), Germany. 
  This work was funded by the Lighthouse Initiative Geriatronics by StMWi Bayern (Project X, grant 5140951), LongLeif GaPa gGmbH (Project Y, grant 5140953), “Centre for Tactile Internet with Human-in-the-Loop” (CeTI, grant 390696704)  
  and
  KI.FABRIK Bayern (grant DIK0249).
S. Haddadin has a potential conflict of interest as shareholder of Franka Emika GmbH.
  }%
  \thanks{$^{2}$The authors are with the Department of Mechanical Engineering at Stony Brook University (SBU), Stony Brook, USA. The work was partially funded by the grant NSF CMMI 1853454, and a SBU OVPR seed grant.} %
}

\makeatletter
\renewcommand{\section}{\@startsection{section}{1}{\z@}{0.35ex plus 0.35ex minus 0.5ex}%
	{0.35ex plus 0.35ex minus 0ex}{\normalfont\normalsize\centering\scshape}}%
\renewcommand{\subsection}{\@startsection{subsection}{2}{\z@}{0.25ex plus 0.25ex minus 0.5ex}%
	{0.25ex plus 0.25ex minus 0ex}{\normalfont\normalsize\itshape}}%
\makeatother

\renewcommand{\baselinestretch}{0.89}
\expandafter\def\expandafter\normalsize\expandafter{%
\normalsize
\setlength\abovedisplayskip{2pt}
\setlength\belowdisplayskip{3pt}
\setlength\abovedisplayshortskip{2pt}
\setlength\belowdisplayshortskip{3pt}
}

\begin{document}

\maketitle
\begin{abstract}
As collaborative robots move closer to human environments, 
motion generation and  reactive planning strategies that 
{allow for elaborate task execution with minimal easy-to-implement guidance whilst coping with changes in the environment} 
is of paramount importance. In this paper, we present a novel approach for generating real-time motion plans for point-to-point tasks using a single successful human demonstration. Our approach is based on screw linear interpolation, which allows us to respect the underlying geometric constraints that characterize the task and are implicitly present in the demonstration. We also integrate an original reactive collision avoidance approach with our planner. We present extensive experimental results to demonstrate that with our approach, by using a single demonstration of moving one block, we can generate motion plans for complex tasks like stacking multiple blocks (in a dynamic environment). Analogous generalization abilities are also shown for tasks like pouring and loading shelves. For the pouring task, we also show that a demonstration given for one-armed pouring can be used for planning pouring with a dual-armed manipulator of different kinematic structure. 
\end{abstract}
\section{Introduction}
\label{sec:intro}

\begin{figure}[t]
\centering
\includegraphics[width=0.99\columnwidth]{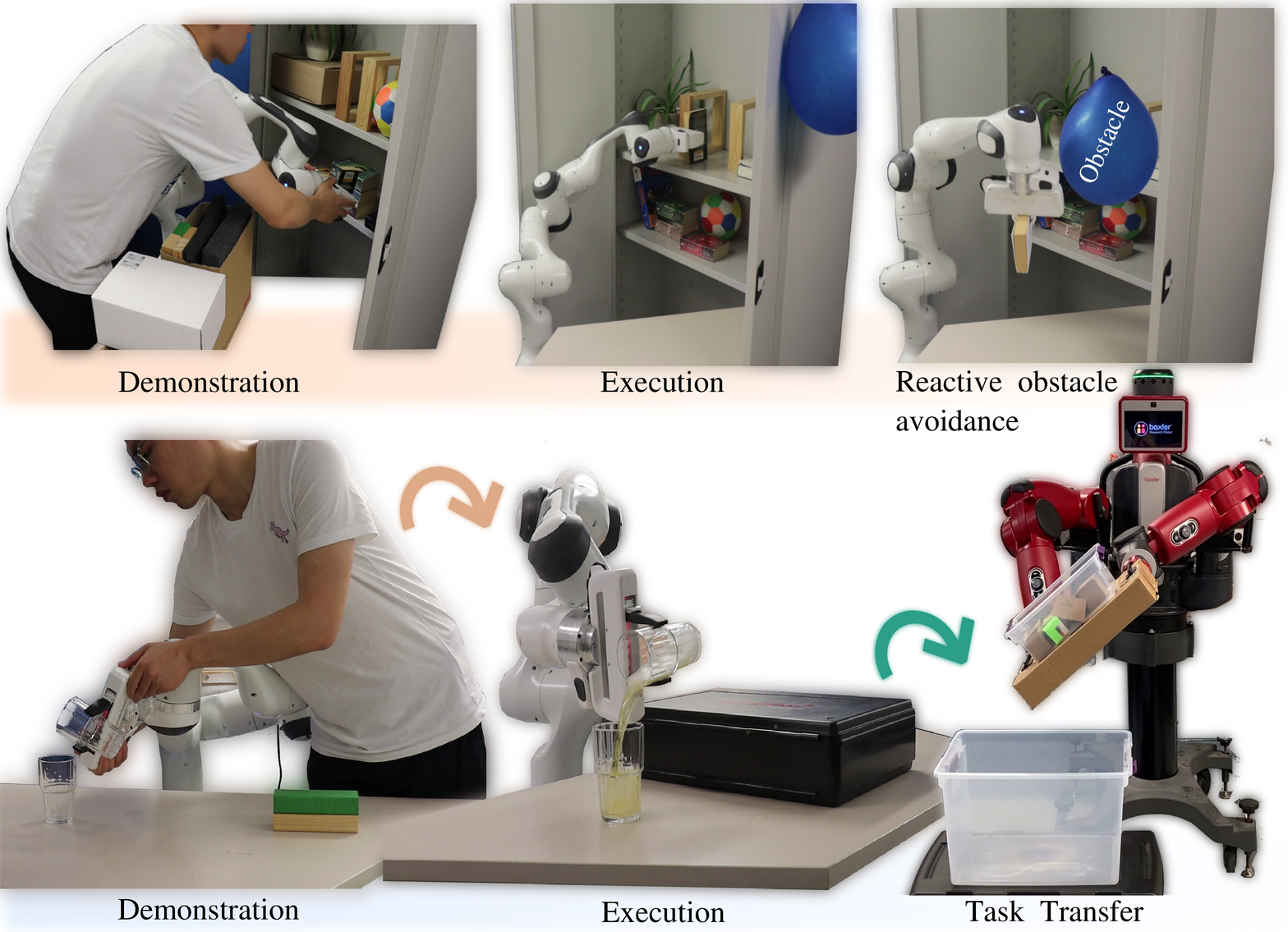}
\caption{Overview of our framework with a shelving task with a single-demonstration and executed in clutter with reactive response to obstacles (top figures), and the generalization of the pouring task towards different placements with collision avoidance and task transfer from a single-arm Panda to a dual-arm Baxter. 
}
\label{fig:tasks_front_page}
\end{figure}

We are interested in the problem of   
a robot manipulating objects in highly constrained,   
unstructured, possibly cluttered and dynamic environments, in a real-time fashion, and the transfer of the task expertise and constraints to different robotic systems.    
Take for instance, the shelving task in Fig.~\ref{fig:tasks_front_page} where the robot needs to generate a sequence of plans to grasp books from a user and place them in different available spaces in the cupboard while avoiding collision with the cluttered cupboard and dynamic objects in the scene. 
Designing plans for elaborate tasks such as this requires experienced roboticists to conceptualize and preprogram the robot \cite{kragic, laha21cooperative}. 
It is even more time consuming and limiting when we want to 
integrate safety aspects, make changes in the environment, structure and/or transfer the task(s) to different robots as seen Fig.~\ref{fig:tasks_front_page}. 

A convenient solution is to teach the robot from human demonstrations. 
However, existing solutions require multiple demonstrations---which can be cumbersome in cluttered and constrained scenarios---and often have limited capability to generalize to different structure and robots. 
Furthermore, practical applicability in industry and service robotics,  calls for the development of a framework which requires minimal time from demonstration (preferably single) to deployment on the robot \cite{Laha_Thesis}. 
In this work, we propose a reactive planner guided by a single-demonstration that enables the robot to solve such elaborate tasks and that can be straightforwardly transferred to 
even multiple-arm systems. 

These are the key problems that our planning framework solves: 
First, our planner produces paths through screw linear interpolation (ScLERP) that implicitly maps---and therefore satisfies---all geometric constraints embedded in the single-demonstration.  
For instance, while pouring a glass of water, the orientation transformations related to the angle deflection along the axis of rotation is coupled with the specific translation position. Our planner explores this information allowing autonomous transfer to new conditions, scenes, new initial or final pose, and even to a completely different robotic system. Note, the user requires no knowledge about the constraints as they are embedded in the topology. 
Second, our planner shifts interpolated points that can lead to collision through a novel reactive approach ensuring feasibility of the motion. 
Finally, our planner provides motion generation in real-time with an additional safety layer for guaranteed collision avoidance within the low-level controller. 

To the best of our knowledge, this is the first work to integrate this level of reactiveness, collision avoidance and generalization in real-time from a single user demonstration. Furthermore, 
we demonstrate through real robot experiments the efficiency of our approach and that the coordinate invariant motions generated by our algorithm can  be  seamlessly  transferred  to  multiple  robots  with different kinematic structure.








\subsection{Related Work}
The basic approach to make robots autonomous for a particular task is hand-engineering of a controller. Yet the generalization of this task across the workspace (different instances) is extremely hard \cite{kragic}. A well known strategy, in this regard, is learning from demonstrations (LfD)~\cite{ravichandar2020recent,calinon2018learning} which is often acquired through teleoperation, kinesthetic or passive observation.
A convenient way to encode the trajectory information is by means of motion primitives, e.g., DMPs \cite{ijspeert2013dynamical,Ijspeert2002}. However, adapting DMPs to reach via-points is not straightforward \cite{paraschos2013promp} and they need multiple demonstrations to encode the mean trajectory which 
can be challenging for rotation.  
Taking a probabilistic approach such as ProMPs improves generalization \cite{paraschos2013promp}, yet their convergence is still only guaranteed within the demonstration region. 
Furthermore, often constraints are embedded in joint-space \cite{Silverio2017,silverio2017learning}, which limits applicability to variations of the joint to task space mapping, e.g., single changes in the kinematics to changing the robot. 
Moreover, most of existing approaches 
have limited generalization capability as well as reactiveness in the presence of dynamic obstacles, and they often  
require a reasonable amount of  demonstrations,  which is cumbersome in highly constrained scenarios. 
%
To avoid multiple demonstrations, 
recent focus has been given to one-shot learning or learning from a single example \cite{WuD10,Zeestraten17RAL,YuFX+18}. Nevertheless, studies in this area often assume 
the existence of an optimization criterion that defines the tasks~\cite{atkeson_schaal_1997} or make strong distributional assumptions about the means of task motion generation~\cite{YuFX+18}. Also, previous works in animation and robotics have incorporated user
inputs in the configuration space \cite{praveena2019user, denny2016theory, phillips1990interactive, gleicher1998retargetting, islam2018online} {where handling obstacles in a reactive fashion is not usually the case \cite{pham2013kinodynamic}}. Our method on the other hand introduces user kinesthetic guidance in task-space planning.


Overall, 
we take a fundamentally different approach.  We explore the one-time kinesthetic demonstration not to learn the trajectory, imitate the human motion, learn a cost function or a policy, but rather to encode and follow the implicit geometric features underlying the path. Our approach is hinged on the observation that from a given set $\mathfrak{S}\subseteq\text{Spin}(3)\ltimes\mathbb{R}^{3},$ the set of all rigid body motions, proper subsets can be drawn from geometric structures of interest which are then implicitly captured by the human demonstration and preserved by ScLERP. In other words, we differ from the literature in the sense that our algorithms embed the constraints in the topology of the rigid-body transformations rather than learning the region of attraction of the dynamical system approximation \cite{ijspeert2013dynamical, nakanishi2003learning}.   
%
In this work, we also exemplify the use of one single demonstration to different instances of tasks, constraints, topologies, and even kinematic chains (such as single-arm to multi-arm system). Also, the resulting deterministic path is safe in the sense that it always mimics part of the demonstration thereby producing paths which are intuitive and predictable for human co-workers.

\section{Problem Formulation}

This section presents the definitions and fundamentals of the planning problem. 
We first briefly recap core concepts regarding the algebra of dual quaternions (DQ) and geometric first-order interpolation properties using screw linear theory. These are the backbone of the proposed approach and
rely on unit dual quaternion 
algebraic and geometric properties. Computational advantages include, e.g., 
 Riemannian 
geometry, 
translation and orientation coupling \cite{Adorno2017,2011_Adorno_THESIS,2005_Wu_Hu_Xu_Li_Lian_TAES,2016_Figueredo_PhDThesis}, 
singularity-free representation,   
 with global convergence controllers   \cite{kussaba2017hybrid,magro2017dual}, 
that embeds 
wrenches, twists and primitives, 
being the universal cover of $SE(3)$ with 
a simply-connected topology (in contrast to $SE(3)$),  
and higher efficiency \cite{1990_Funda_Paul_TRA,OZGUR201666,2019_Xialong_etal__JIRS__DQefficiency_invDyn_Parallel,1998_Aspragathos_Dimitros_TSMC}. 

\subsection{Problem and Mathematical Background}


Our aim is to explore, adapt or define a path, which consists of a sequence of rigid-body transformations,  to complete a geometric constrained task---a motion planning problem. 
%
To connect the spatial transformations and generate a smooth curve in $\SE{3}$, 
we must (i) define the rigid body transformation in the Lie group of   $\unitdualquatgroup$, 
(ii) describe its Riemannian geometry and geodesics, 
and finally (iii) define the screw linear interpolation based on the  geodesic direction. 



An arbitrary\textbf{ rigid body transformation}
%
%
can be represented by the unit dual quaternion $\dq x \in \unitdualquatgroup$, 
\begin{equation}
    \dq x=\quat r+\tfrac{1}{2}\dual\quat p\quat r,  
    \label{eq:dq}
\end{equation}
where  
$\quat r=\cos(\phi/2)+\sin(\phi/2)\quat n $   
represents a rotation with angle $\phi$ around the axis $\quat n$, in unit quaternions $\unitquatgroup$   
\cite{kuipers:1999}, 
$\quat p$ is a pure quaternion 
that represents the translation, and 
$\dual$ is such that 
$\dual{\neq}0$ but $\dual^{2} {=}0$, \cite{Selig2005}.  
Unit dual-quaternion, $ \unitdualquatgroup $, is a Lie group with inverse element being  $\dq x ^\ast =\quat r ^\ast  +\frac{1}{2}\dual \quat r ^\ast  \quat  p ^\ast $, and identity $\dq{1}$.     
%
Dual quaternion elements can also be described by $\dq x=\getp{\dq x}+\dual\getd{\dq x},$
where $\getp{\dq x}$ and $\getd{\dq x}$ are the primary and dual components. 

From its differentiable \textbf{Riemannian geometry}, 
%
%
it is endowed with a collection of inner products on the tangent space at $ \unitdualquatgroup$, which in turn builds a Riemannian metric \cite{Book:Boothby:2002,2017_Busam_Birdal_Navab_ICCVW}. 
Once a Riemannian metric is assigned to the manifold---e.g., the length of the path \cite{1995_Park_JMD_ASME}---we explore the 
minimum curve length
, i.e.,  
the geodesics 
\cite{2017_Busam_Birdal_Navab_ICCVW}, see \cite{1995_Park_JMD_ASME,Book:Boothby:2002,Zacur2014b,2009_Sachkov_JMS}. 
%
%
%
In such manifolds, 
actions in the geodesics can be expressed by means of the exponential map 
$\exp _{\dq x} : \mathcal{T}_{\dq x} \unitdualquatgroup  \rightarrow \unitdualquatgroup  $. 
The $\exp _{\dq x} $ locally maps a vector in the tangent space   
$\mathcal{T}_{\dq x}  \unitdualquatgroup $ (at  $\dq x  \in  \unitdualquatgroup$) 
to a point on the manifold following the geodesic through $\dq x $,  \cite{BOOK:2016:Gallier_Quaintance,2017_Busam_Birdal_Navab__ArXiv}.  
The inverse mapping (from manifold to tangent space at the point $\dq x$) is the logarithm map 
$ \log _{\dq x} : \unitdualquatgroup \rightarrow  \mathcal{T}_{\dq x} \unitdualquatgroup  $. 

The mappings  
$\exp _{\dq x} $  
and 
$\log _{\dq x}$  
are non-trivial to obtain. A solution is to compute them by parallel transport \cite{BOOK:2016:Gallier_Quaintance,2017_Busam_Birdal_Navab_ICCVW,2013_Lorenzi_Pennec__IJCV}. The parallel transport exploits   
 the exponential function that maps vectors from the tangent space (at the identity) to the manifold \cite{Zacur2014b}, 
 \begin{align}
     \exp _{\dq x}(\dq y)   & = \dq x \exp ( \dq x^\ast \dq  y  ),    \nonumber \\
     \log _{\dq x}(\dq z)   &  = \dq x \log ( \dq x^\ast \dq z  ) ,   
     \label{eq:parallelTransport}
 \end{align}
where  $\dq  z \in \unitdualquatgroup  $ and $\dq  y $  is defined in the tangent space at $\dq x$---notice that $\dq y$ is not an unit DQ. 
%
%
The  
$\exp$ and $\log$  maps from the tangent space, at the identity, i.e., $\mathcal{T}_{\dq 1} \unitdualquatgroup$ are given by the dual vector representing the axis of screw motion and the dual angle containing both the translation length and the angle of rotation, see further details in  \cite{2017_Busam_Birdal_Navab_ICCVW,2017_Busam_Birdal_Navab__ArXiv,sarker2020screw,2012_Wang_Han_Yu_Zheng_JMAA}.  

Finally, to describe the \textbf{screw linear interpolation} that connects two points   $\dq x_{1}$ and $\dq x_{2}$, 
and to find points in the path 
 $\dq x(\tau):[0,1]\rightarrow\unitdualquatgroup$
 with $\dq x(0)=\dq x_{1}$ and
$\dq x(1)=\dq x_{2}$, 
we exploit 
\eqref{eq:parallelTransport}. 
%
%
First, 
we map 
$\dq x_{2}$ following the geodesic on $\unitdualquatgroup$ through 
$\dq x_{1}$ onto the tangent space 
at $\dq x_{1}$.
Naturally, this mapping yields a vector in the $\mathcal{T}_{\dq x _1} \unitdualquatgroup$
corresponding to the geodesic direction of $\dq x_{2}$ with respect
to $\dq x_{1}$. Hence, 
\begin{equation}
    \log_{\dq x_{1}}(\dq x_{2})=\dq x_{1}\log(\dq x_{1}^{\ast}\dq x_{2}),    
    \label{eq:parallel_for_x1_to_x2}
\end{equation}
where 
$\log_{\dq x_{1}}$ is computed
using parallel transport.
%
Notice that \eqref{eq:parallel_for_x1_to_x2} is defined in the tangent space of a Riemannian manifold, hence it is a vector space with basis defined  by a vector field. Thus, we can linearly interpolate points and compute any point between along the geodesic direction starting from $\log_{\dq x_{1}}(\dq x_{1})$ towards $\log_{\dq x_{1}}(\dq x_{2})$, as 
$
    (\log_{\dq x_{1}}(\dq x_{2})-\log_{\dq x_{1}}(\dq x_{1}))\tau +\log_{\dq x_{1}}(\dq x_{1}). 
$
Note, however that  $\log_{\dq x_{1}}(\dq x_{1})=0$.  
Hence, 
%
%
using parallel transport to map the vector  in $\mathcal T _{\dq x_{1}}{\unitdualquatgroup}$
back to the unit DQ manifold following the geodesics
through $\dq x_{1}$ yields
\begin{align}
\dq x(\tau)= & \exp_{\dq x_{1}}\left(\dq x_{1}\log(\dq x_{1}^{\ast}\dq x_{2})   \tau \right)\nonumber \\
= & \dq x_{1}\exp\left(\log(\dq x_{1}^{\ast}\dq x_{2})   \tau  \right).\label{eq:sclerp_equation}
\end{align}

\begin{lem}[ScLERP -- \cite{DQBlending}]\label{lem:sclerp}
Given two rigid body transformations, $\dq x_1 $ and $ \dq x _2$, the ScLERP function,  
\begin{equation}
    \sclerp{\tau,\dq x_1,\dq x_2} 
    = \dq x_{1} (\dq x_{1}^{\ast}\dq x_{2})^\tau, 
    \label{eq:sclerp function}
\end{equation}
returns any rigid pose transformation along the geodesic direction from  $\dq x_1 $ to $ \dq x _2$ scaled linearly along $\tau \in [0,1]$. Taking equally spaced values within $\tau$ yields, therefore, a  
screw linear interpolation from  $\dq x_1 $ to $ \dq x _2$.  
\end{lem}

Notice the ScLERP function \eqref{eq:sclerp function} is the same as the one derived in \eqref{eq:sclerp_equation}. This can be shown by geometrical exponential \cite{Adorno2017,article:1996_Kim_Kim_Shin__JVCA},  
and from the scaling of the dual rotation angle about the screw axis---hence the name \cite{DQBlending}. 
%

\begin{remrk}
The ScLERP interpolation explores the natural parametrization of screw coordinates in terms of $6$-DoF displacements \cite{Allmendinger2018, sarker2020screw}. They are particular attractive for coordinate-invariant interpolation which is not possible when decoupling orientation and translation \cite{grassmann2018smooth}, as detailed in \cite{Allmendinger2018}.
Similar interpolation scheme nonetheless could also be derived from  $\SE{3}$, 
and other covering groups that satisfy left-invariance and are based on non-minimal representation of rigid displacements.  
Hence, it is by no means restricted to the choice of $\unitdualquatgroup$. 
Still, a matrix-based solution is non-attractive due to the additional computational cost---that can possibly restrict real-time implementation---and due to the efficiency, compactness and intuitiveness of $\unitdualquatgroup$ which can depict wrenches, twists, geometric primitives, constraints and its tangent space with the same algebra.   
\end{remrk}

\subsection{Overview of Problem}

In this work, we are interested in the motion planning problem  
that completes a geometric constrained task reactively in a real-time fashion, while responding to unforeseen events, such as dynamic obstacles with guaranteed avoidance behaviour---also in real-time. 
%
 
The proposed motion generation scheme takes as prior knowledge a single successful task demonstration---defined by a sequence of poses expressed as unit dual quaternion
    \begin{equation}
        \mathcal{DP} = \{{\dq d}_1, {\dq d}_2, \cdots, {\dq d}_n \}, ~\dq d_i \in \unitdualquatgroup
        \label{eq:demonstrations}
    \end{equation}     
Since the task demonstration is a successful one, it satisfies the task-constraints, and this knowledge is implicitly embedded in the demonstration. As an example, during pouring, the axis and angle of rotation is implicitly present in the demonstration. We can obtain the demonstration, $\mathcal{DP}$, either by directly sensing the end-effector pose or by recording the joint-space path from the joint encoders and using the forward kinematics map for the manipulator. Kinesthetic teaching along with joint space path recording is used in this paper.  
A key point here is that no other information, or knowledge, about the task or its constraints are required or needs to be provided. 

Given such implicit constraints, our planner aims at finding a sequence of rigid body transformations---herein named the final path and the corresponding robot joint-space actions  
that takes the manipulator end-effector from 
an initial configuration $\dq x _0$ to a final goal $\dq x _f$ while satisfying the constraints observed in
$    \mathcal{DP} $. 
Formally, our problem can be defined as follows:

\noindent \textbf{Problem Definition: }
Given a single user-demonstrated path $\mathcal{DP}$, and the new initial and final pose $ \dq x _0$ and $ \dq x _f$, respectively, find a path from $\dq x_0 $ to $\dq x_f$ such that \begin{enumerate}
    \item All constraints that are implicit in $\mathcal{DP}$ are satisfied;
    \item Obstacles that prevent motion feasibility  are avoided; 
    \item Motion generation is achieved in real-time with an additional safety-layer for low-level controller guaranteeing collision avoidance.
\end{enumerate} 

\section{Task Space Imitation Algorithm (TSIA)}
\label{sec:user_guided_traj_generation}


In this section, we present a solution to satisfy the first condition in the problem definition. 

\noindent \textbf{Problem Statement 1:} \textit{Compute a path in $\text{Spin}(3)\ltimes\mathbb{R}^{3}$ starting at an initial pose $\dq{x}_{0}$ and ending at a goal pose $\dq{x}_{f}$ 
while still maintaining the task relevant motion constraints implicit in the user demonstration.}


%
From the human demonstration we obtain a series of unit dual quaternions that encode the rigid body constraints along the path, i.e., 
$ \mathcal{DP} $ as in \eqref{eq:demonstrations}, where  
    %
    ${\dq d}_1$ corresponds to the starting pose of the demonstration and ${\dq d}_n$, the final pose.

    Let ${\dq d}'_n $ ($=\dq{x}_{f}$) be the new goal pose.
    %
    To ensure that the relative transformations between the poses in the demonstrated path are preserved on the path to the new goal we want to reach, we replicate the demonstrated motion with respect to the new goal and call this path the \textit{imitated path} ($\mathcal{IP}$). The $\mathcal{IP} $ reflects the constraints present in the demonstration. 
    
    To compute $\mathcal{IP}$, we first obtain the transformation, $\dq{\delta}_i$, between the last pose on the demonstrated motion ($\dq d_n {\in} \mathcal{DP}$) and every other pose on the demonstrated motion, i.e., 
\begin{equation}
\label{equ:transformation_imi}
\dq{\delta}_i =  {{\dq d}^*_{i-1}}  {\dq d}_n  ,\ i = 2, \dots, n.
\end{equation}
Then, the imitated path $\mathcal{IP} = \{{\dq d}'_1, {\dq d}'_2, \cdots, {\dq d}'_n\}$ can be calculated using, 
\begin{equation}
\label{equ:imitated_path}
{{\dq d}'}_{i-1} = {{\dq d}'}_{n}    {\dq{\delta}_i}^*  ,\ i = 2, \dots, n.
\end{equation}
%
%



\noindent \textbf{Final Path: }The calculation of the final path $\mathcal{FP} = \{{\dq d}''_1, {\dq d}''_2, \cdots, {\dq d}'_n\}$ from a new initial pose ${\dq d}''_1$ ($=\dq{x}_{0}$) to the new goal pose ${\dq d}'_n$ is performed using ScLERP~\cite{laha21p2p}.
Intuitively speaking, our goal is to find a path that blends into the $\mathcal{IP}$ and after blending just follows the $\mathcal{IP}$ to obtain the same geometric constraints during motion as in the $\mathcal{DP}$. 

Let ${ \dq d}'_i$ be a {\em guiding pose} (in this regard there are various possibilities to explore depending on the scenario) on the $\mathcal{IP}$ and ${ \dq d}_c$ be the current task space pose of the manipulator---starting from $\dq x_0$. We now compute a target pose $\dq d_t$ using ScLERP in dual quaternion space, that is, 
\begin{equation}
\label{eq:interp}
{\dq d}_t(\tau) = {\dq d}_c \cdot \left( {\dq d}_c^{*} \cdot {\dq d}_i' \right)^{\tau}{,}
\end{equation}
where $\tau \in [0, 1]$ is the time primitive. Different choices of the parameter $\tau$ give different target poses in task space. Note that $\tau=0$ corresponds to ${\dq d}_c$ and $\tau=1$ corresponds to ${\dq d}_i'$. We then select an interpolation pose\footnote{As an heuristic, we suggest deploying guiding poses within $20 \% $ to $70 \%$ of the $\mathcal{IP} $  length.} and use that pose as a reference ($\dq x_d$) for our kinematic controller. With the resulting joint velocities $\dot{\myvec q}$---from the closed-loop---the robot takes a new configuration closer to $\dq x_d$ which is thereafter set as the new current pose. The process continues taking ${\dq d}_{i{+}1}'$ as the new guiding pose on $\mathcal{IP}$.
This follows until the goal is reached (we define the error $\bm{e}({ \dq d}_c,{\dq d}'_n) = \norm{\vector{({ \dq d}_c),\vector{({\dq d}'_n)}}} >$ a tolerance). 
\begin{algorithm}[t]
		\caption{Reactive Imitation Algorithm}
        \label{alg:1}
		\begin{algorithmic}[1]
		\Procedure{Get Final Path:}{${\dq d}'_n$, $\mathcal{DP}, \mathcal{O}$}
%
            \State  $\dq{\delta}_i \gets$ Compute spatial difference with $\mathcal{DP}$ \eqref{equ:transformation_imi};
            \State  $\mathcal{IP} \gets $ Get imitated path with $(\dq{\delta}_i, {\dq d}'_n)$ \eqref{equ:imitated_path};
            \While{$\bm{e}({ \dq d}_c,{\dq d}'_n) >$tol and $i \leq n$}
                \State$ { \dq d}_c \gets$ Get current pose;
                \State ${ \myvec{o}_c } \gets$ Compute closest obstacle point (${ \dq d}_c, \mathcal{O}$);                       
                    \If{inside detection shell  ($ { \dq d}_c, \myvec{o}_c$) }
                        \State $\bm{T}$ $\gets$ Escape Tree Generation (${ \dq d}_{\textrm{cnew}}$,  $ \myvec{o}_c$);
                        \State ${ \dq d}_{c} \gets$ Set new current pose ($\bm{T}$);
                    \EndIf                    
                \State${ \dq d}'_i \gets$ Set guiding pose;
                \State$\dq d_t \gets$ ScLERP (${{ \dq d}_c, \dq d}'_i$); 
                \State $\dot{\myvec q} \gets$ Compute reactive control action ($\dq d_t$) \eqref{eq:controller};
                \State $i \gets i + 1$;
            \EndWhile
\EndProcedure
	\end{algorithmic}
	\end{algorithm}
Building our user-guided motion generation approach based on ScLERP is crucial in order to ensure constraint satisfaction during blending, i.e., during approach through interpolation to the desired imitated path. Since ScLERP is based on exponential and logarithmic maps (see \eqref{eq:parallelTransport}), 
it is clear that translation and orientation of interpolated poses between key-points remain bounded by both boundary pose. Hence, if both $\dq d'_i$ and $\dq d'_{i{+}1}$ satisfy a constraint, then the interpolated motion also concurs. 
Note that any projected subset can also make use of the exponential and logarithm mapping and constraints would also be satisfied. 
Hence, it is a property intrinsic of our method to implicitly satisfy all geometric constrains defined in the task demonstration without having to explicitly define it---which would require a roboticist with good expertise and geometric reasoning to design. 

Finally, note that such constraints cannot be guaranteed without a proper screw linear interpolation. A simple example is the usual approach based on decoupling translation and attitude. Even when using proper attitude spherical interpolation, different key points in a  rigid body leads to different unpredicted trajectories, see details in \cite{Allmendinger2018}. 

   \begin{remrk}
    Note that we are not learning the trajectory itself, 
    but rather we are automatically and implicitly transferring the rigid body displacements which naturally embeds the desired task constraints. This facilitates generalization to different initial and end-goals, topologies, slightly different tasks, and even to different kinematic-chains, such as from single-arm to dual-arm---as seen in Section \ref{sec:simulations&exp}.  
    \end{remrk}
\section{Reactive Obstacle Avoidance}
\label{sec:obstacle_avoidance}

During the real-time motion generation, unforeseen events 
such as the presence of obstacles 
may make the path planned by TSIA infeasible. 
We now present a reactive approach for real-time obstacle avoidance that we use to modify the path generated by TSIA and hence ensure task success. More formally, we present a solution to the following problem:
%
 
\noindent \textbf{Problem Statement 2:} \textit{Compute a collision-free path in $\text{Spin}(3)\ltimes\mathbb{R}^{3}$ while still respecting the constraints embedded in the path in the last step.}\\
For didactic reasons, let us consider 
a \textit{sphere of influence} around an obstacle with a radius larger than the obstacle. 
The robot path is deflected only when it is inside this detection shell. To enable integrated real-time collision-free path during planning we introduce the Rapidly Expanding Plane-oriented Escaping Trees (REPET) algorithm.      
%

\noindent \textbf{Escape Tree Generation: }
As the robot moves along the real-time generated path from the current pose  $\underline{\bm{x}}_{c}$ 
towards the detection shell of the closest obstacle, centred at $O_{\text{obs}}$, we start devising the avoidance scheme.  
%
First, we obtain the normal vector ${\bm{\eta}}(\trans{ \dq x _c }, O_{\text{obs}})$,  
given by ${\bm{\eta}} : \realset^3 {\times} \realset^3 \rightarrow \realset^3 $ based on the current position $\trans{ \dq x _c }: \unitdualquatgroup \rightarrow \realset^3 $ and the closest point to the surface of the obstacle described by the shell centre.   
%
%
%
%
Then, a tangent plane $\bm{p}_{\text{t}}$ is computed using the normal information, which is thereafter used to build  
%
a vector ${\bm{v}}$ orthogonal to ${\bm{\eta}}$ ({e.g., using unit vector $\hat{i}$}), 
\begin{equation}
    \bm{v} =  \hat{i} \times \frac{ \bm{\eta} }{ \norm{\bm{\eta}} } \cdot \Big( \frac{k_\eta}{2}\Big),
    \label{equ:plane_ortho}
\end{equation}
where $k_\eta$ denotes the length of the diagonal of the plane. 
In the same fashion, we define ${\bm{u}} = \getp{\bm{p}_{\text{t}}} \times \bm{v}$ which is used along with $\bm{v}$ to obtain points on sides and edges of the plane.

For efficient collision avoidance, 
we seek to explore poses
along the normal vector in the tangent space of the surface---ensuring guaranteed collision free points surrounding convex obstacles or good exploration along non-convex ones.  Herein, for brevity, we are focusing on convex ones.  
The core idea is to integrate such key avoidance points into our motion generation scheme, which is  based on a sequence of transformations interpolated through ScLERP. 
Hence, the key points shift\footnote{The deflection along the path is deployed in Cartesian coordinates only as we want to disturb the least the constraints. Still, the integrated path remains connected through ScLERP ensuring the remaining constraints are satisfied during avoidance.} a sequence of transformations within the $\mathcal{FP}$ building a collision-free path along the tangent space of the closest obstacle's detection shell. 

As shown in Alg.~\ref{alg:2}, we first set a root (based on current pose $\underline{\bm{x}}_{c}$) in the tangent plane, and keep exploring vectors along the orthogonal plane \eqref{equ:plane_ortho} with distance $\kappa _\eta$ from the shell intersection point for fast building of subsequent planes. The length of $\kappa _\eta$ defines the avoidance strategy. Large $\kappa _\eta$ ensures good avoidance but larger deviations from the path, while smaller $\kappa _\eta$ provide smoother motions yet may not be enough to avoid within one single step. Next, we use the rapid plane generation scheme to grow multiple planes iteratively, thus generating numerous paths exploring different avoidance possibilities which we check for {end-effector} collisions at each step. This leads to a tree that grows along the new plane-key-points generation, which we call the \textit{escape tree}.

\begin{figure}[t]
\centering
\includegraphics[scale=0.26]{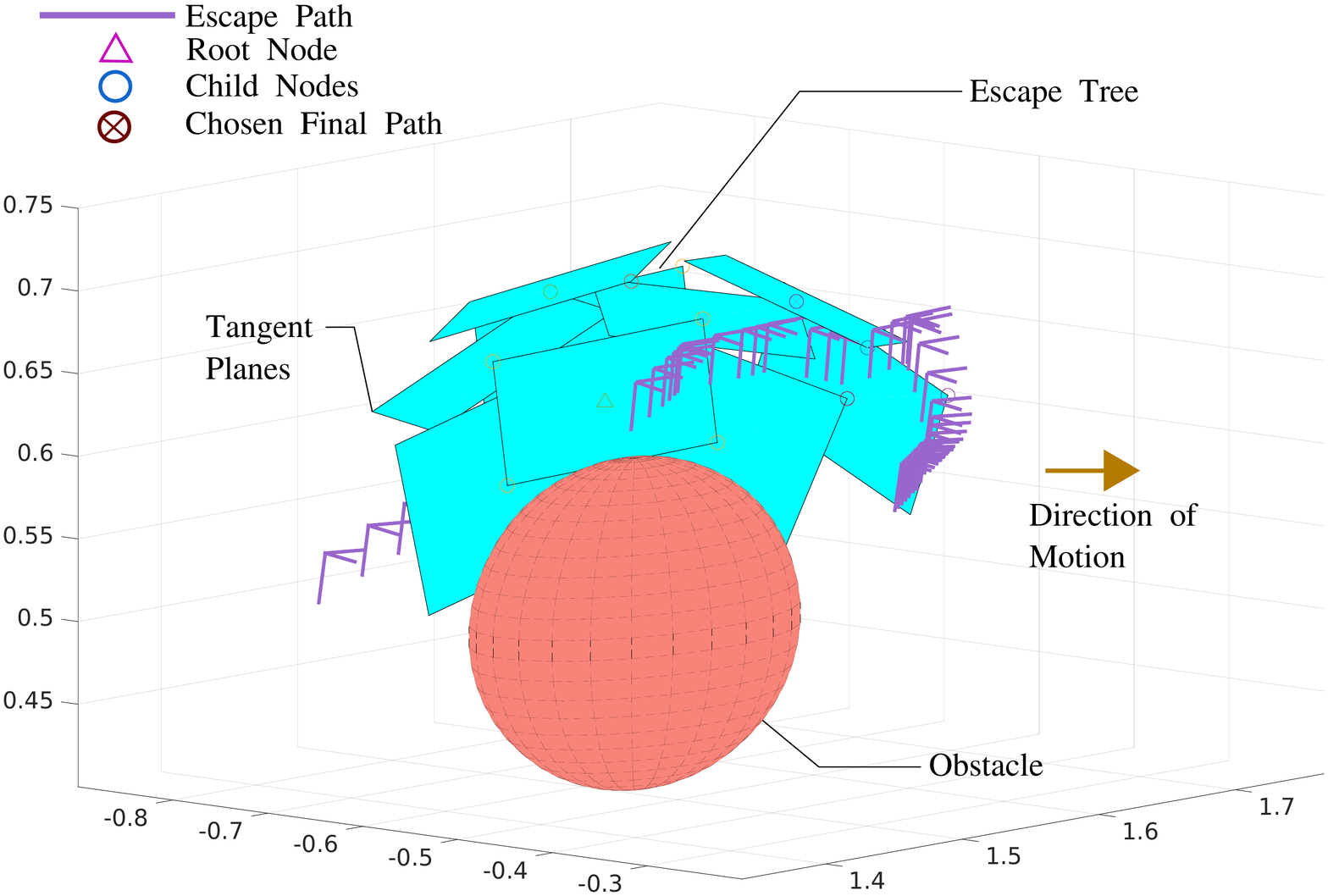}
\caption{Fast Obstacle avoidance using REPET showing all the tangent planes. This collision avoidance path is generated for the stacking problem.}
\label{fig:escape_tree}
\end{figure}

The tree is expanded until either a limited level is reached (stop criterion) or one leaf finds a free Cartesian path towards a free pose within the final path---located after the obstacle shell (exception being if the final goal is within). 
If we have multiple free path leafs in the breadth, we select the optimal one $\bm{\varrho}$ in terms of a heuristic cost function. Herein, we are taking the points closer to the goal. 
If the limited level is reached, we sample random new points in the plane with larger $k _\eta$.  The resulting path follows the tree with our TSIA path generation scheme with new key avoidance poses. 
An example of REPET is shown in Fig.~\ref{fig:escape_tree} (Alg. \ref{alg:2}) for the real-robot experiment in Section \ref{sec:simulations&exp}.A. Alg.~\ref{alg:1}
illustrates TSIA combined with the reactive collision avoidance. 


\begin{algorithm}[t]
		\caption{REPET Algorithm}
        \label{alg:2}
		\begin{algorithmic}[1]
		\Procedure{Build Plane:}{origin ${\dq x}_{\rm c}$, $k_\eta$, obstacle o}
        \State ${\bm{\eta}} \gets$ Get normal vector (${\dq x}_{\rm c}$, o);
        \State $\bm{v} \gets$ Orthogonal vector defining plane (${\bm{\eta}}, k_\eta$) ~(\ref{equ:plane_ortho});
\EndProcedure
	\Procedure{Escape Tree Generation:}{${\dq x}_{\rm c}$, ${\dq d}'_n$, $\myvec{o}_c$}
	\While{stopping criteria is not satisfied}
	    \State{${\dq x}_{\rm r} \gets$  Set root (${\dq x}_{\rm c}$);}
	    \For{Each leaf $j$ of the decision tree level;}
	        \State{${\dq p}_{\rm c} \gets$  Build plane (${\dq x}_{\rm r}, k_\eta, \bm{o}_c)$;}
	        \State ${\dq c}_{\rm j} \gets$ Get child from root;
	        \State{${\dq {cc}}_{\rm j} \gets$ Check for collisions;}
	        \State ${\dq c}^{*}_{\rm j} \gets$ Select best based on criteria (${\dq {cc}}_{\rm j},{\dq d}'_n$);
%
	       %
	    \EndFor
	\EndWhile
	\State{$\bm{\varrho}, { \dq d}_o \gets$Select path, pose (${\dq c}^{*}_{\rm j}$);}
	\EndProcedure

	\end{algorithmic}
	\end{algorithm}
\section{DQ Based Kinematic Controller}
\label{sec:kine_controller}

As stressed in the problem definition, 
the motion generation needs to be achieved in real-time 
with an additional layer for low-level controller that guarantees collision avoidance.  
%
%
Essentially, the idea is to have a feedback controller with exponential convergence to a desired reference pose in $\text{Spin}(3)\ltimes\mathbb{R}^{3} $ without decoupling the translational and rotational components. First, we can define the spatial difference $\dq x_e$ as,
$
    \dq x_e =\dq x^{*}_m\dq x_d\,,
$
where $\dq x_m$ denotes the measured current pose and $\dq x_d$ the desired pose of the end-effector. The error metric ${\dq{e}}$ can then be defined as ${\dq{e}} = 1 - \dq x_e$. This can be rewritten and mapped back into $\mathbb{R}^8$ using the Hamilton operator $\hami -$ (as in \cite{agrawal1987hamilton,2011_Adorno_THESIS,Adorno2017}) and differentiated as,
\begin{equation}
    \vector\dot{\dq{e}} {=}\hamilton -{\dq x_d}\bm{C}_8 \vector\dot{\dq x}_m\,,
\end{equation}
where $\bm{C}_8$ is a diagonal matrix and $\vector:\mathcal{H}\rightarrow\mathbb{R}^{8} $ as defined in \cite{2013_Figueredo_Adorno_Ishihara_Borges_ICRA,Adorno2017}. Now, replacing $\dot{\dq x}_m$ with the robot Jacobian mapping leads and pseudo-inverse controller leads to  
\begin{equation}\label{eq:controller}
     \dot{\myvec q} = - (  \hamilton -{\dq x_d}\bm{C}_8 \bm{J}    )^{+} \vector\dot{\dq{e}} =  -  \textbf{N}^{+} \lambda_e  \vector{\dq{e}}, 
\end{equation}
where $\dot{\myvec q}$ is the joint velocities, $\textbf{N}^{+}$ is extended Jacobian pseudoinverse and $ \lambda_e $ is a positive gain. This controller shows coordinate left-invariance which matches our framework. Due to space limits, we will omit further details or proofs but they can be found in \cite{2013_Figueredo_Adorno_Ishihara_Borges_ICRA}.  In this work, we also include additional nullspace tasks such as joint-limit avoidance. 
%
%

For ensuring avoidance motion only along the constrained obstacle surface within the low-level controller, we continuously modify the desired pose---as it reaches the obstacle detection shell---by means of 
\begin{equation}
\begin{split}
    {\dq{x}_{d}}_{\text{obs}} = {}&\dq x_m\exp \left(
                \log(\dq x_m^{*}\dq x_d) - \tfrac{1}{2}\trans{ \dq x_m^{*}\dq x_d }\right)\\ &+ \dq v_{\text{ee}}\left(\bm{I} -  \mathcal{P}(\bm{\eta}_{\text{obs}})\right)\vector\left(\tfrac{1}{2}\trans{ \dq x_m^{*}\dq x_d } \right)
\end{split}
\end{equation}
where 
$\trans{ \dq x_m^{*}\dq x_d) }$ is the translation element of the spatial difference,  
$\dq v_{\text{ee}}$ represents the current linear velocity of the end effector and $\bm{\eta}_{\text{obs}}$ denotes the normal vector, viewed from the rigid body frame, of the nearest point on the obstacle surface. $\mathcal{P}(\bm{\eta}_{\text{obs}})$ is the orthogonal projection for which $\bm{\eta}_{\text{obs}}$ belongs to its range space. The last term makes sure that no action will violate the constrained surface, that is, any action along the obstacle normal belongs to the nullspace of the solution.  
Notice that similar strategy has been used before for $SE(3)$ collision avoidance \cite{1995_Park_JMD_ASME,2001_Han_Park_TAC}. However, to the best of the authors' knowledge, this is the first work to extend it to dual quaternion algebra for real-time deployment.

\section{Experiments and Analysis}
\label{sec:simulations&exp}

In this section, we present a set of experiments to 
evaluate our proposed framework. 
We show that our reactive user-guided motion generation scheme  
can successfully generalize tasks to different initial and final conditions while ensuring 
embedded constraints from demonstration are satisfied. Our planner is able to evade unforeseen obstacles\footnote{Herein, we assume full knowledge about  obstacle poses, as detection is out of scope for this work which is agnostic to the detection strategy. 
} in real-time 
while satisfying task constraints.    
Finally, task demonstrations given on one manipulator (Franka Emika Panda)\footnote{Experiments executed at the Technical University of Munich.} can be transferred to generate motion plans for a different robot, Baxter,\footnote{Experiments conducted at Stony Brook University.}  with distinct kinematic structure. Demonstrations given for a one-armed task can also be used to generate a plan for the same task done in a bimanual fashion. 

Experiments were performed with the Panda arm, in the first three tasks: stacking, pouring, and shelving books. For showing generalization across different types of robot arms, we used the pouring demonstration on the Panda arm to generate dual-armed pouring motion plan with the Baxter.
Note that for all experiments, a single demonstration was used, without any need for adjustments or corrections. 

\begin{figure}[t]
\centering
    {\scriptsize
\def\svgwidth{0.95\columnwidth} 
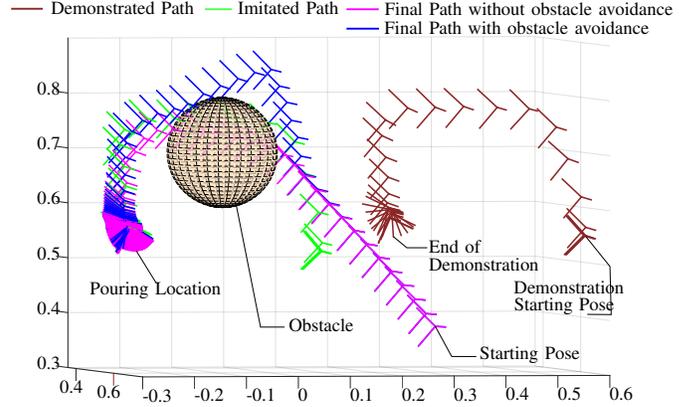
} 
\caption{Paths for the pouring task as seen in Fig.~\ref{fig:tasks_front_page}. Successful generalization for different initial and final conditions without obstacles (magenta curve) and reactive real-time deflection (blue).  
}
\label{fig:task_pouring_path}
\end{figure}%


\begin{figure}[t]
\centering
    {\scriptsize
\vspace{5pt}
\def\svgwidth{0.95\columnwidth} 
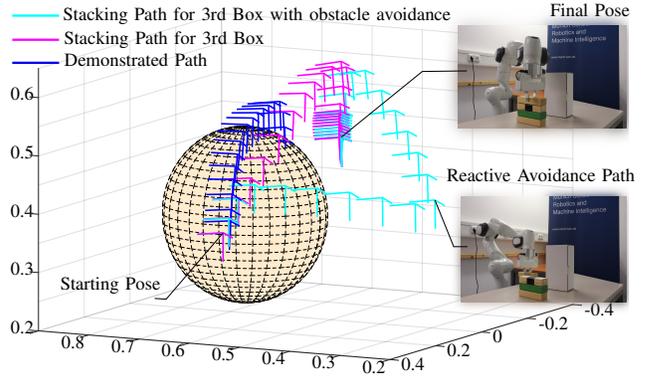
} 
\caption{Demonstrated path (blue), final computed path without collision avoidance (magenta), collision free path (cyan) for the stacking task (Fig.~\ref{fig:shelf_strip}). Note the integration of the pre-grasp path for placing the blocks even in the presence of obstacles.}
\label{fig:task_stacking_paths}
\end{figure}%

\subsection{Analysis and generalization under different conditions}

\begin{figure}[t]
\centering
\includegraphics[scale=0.24]{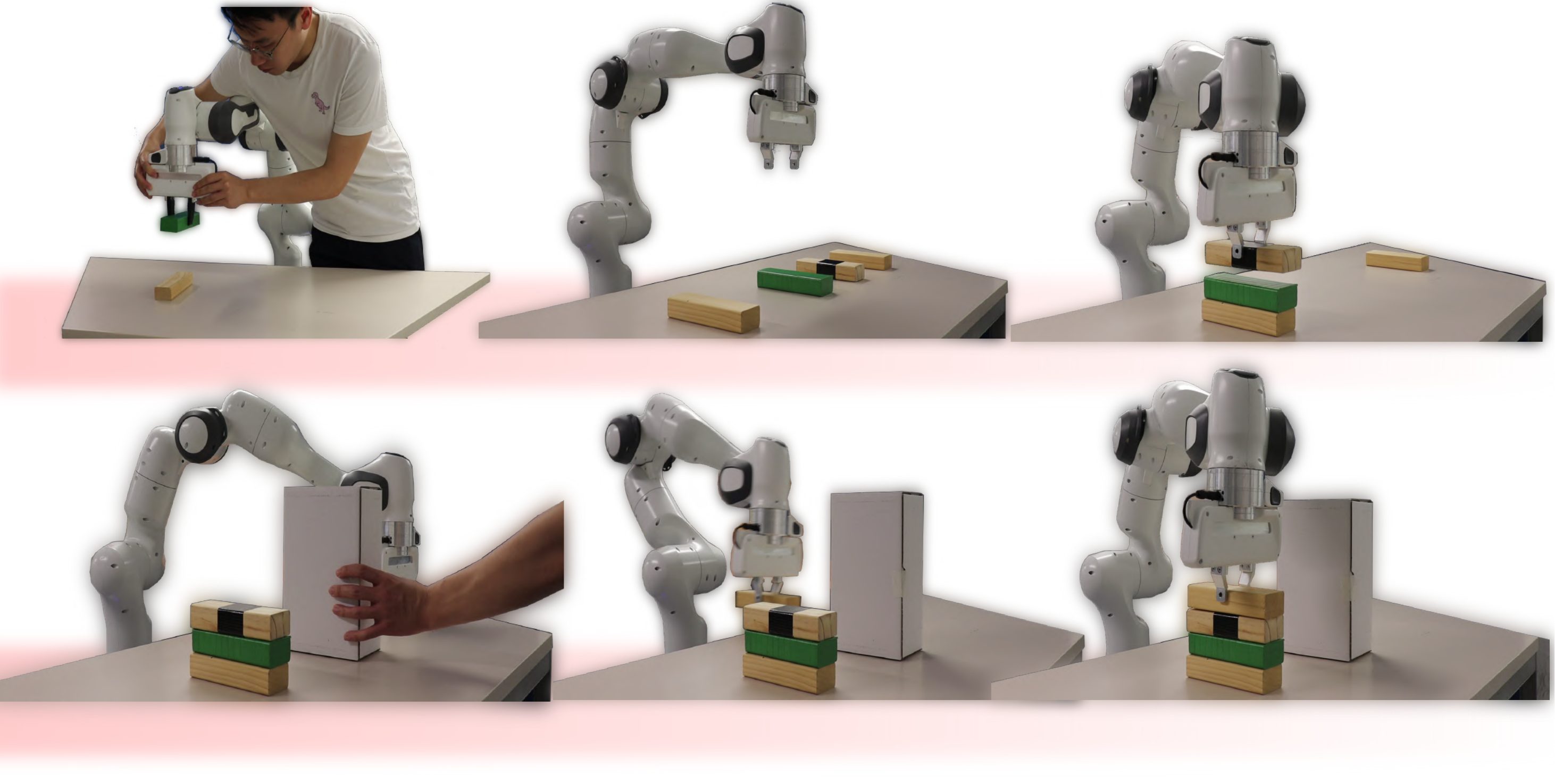}\vspace{-5pt} 
\caption{Overview of our approach from single-demonstration (top-left) to generalization and sequential manipulation to reactive collision avoidance (bottom figures) 
}
\label{fig:shelf_strip}
\end{figure}

First, to better understand and validate the proposed framework, 
we performed two different tasks:\footnote{The guiding pose was set to be $20\%$ of $\mathcal{IP}$ length.}  
(i) Pouring water from one glass to another with shifting positions of the glasses;  
(ii) Stacking rectangular wooden cuboids. 
Both tasks were demonstrated only once, and executed ($\tau=0.01$) under different conditions as shown in Figs.~\ref{fig:tasks_front_page} and \ref{fig:shelf_strip}. 

For pouring, the challenge was to embed the angle transformations around a specific axis of rotation---while executing the task in a completely different location with obstacles in the scene. 
Manually designing the task would be  challenging, yet our algorithm embeds such constraints in the demonstrated path  
which is ensured by the imitated path and deployed for the final path through ScLERP.  
All the trajectories 
are shown in Fig.~\ref{fig:task_pouring_path}, which depicts that our reactive framework ensures feasibility of the task while, at the same time, satisfying the demonstrated constraints.

The stacking task is a sequential manipulation problem, where a single demonstration is given but five sequence of motions, from grasping the first block to  releasing the last bock needs to be executed. The only additional information is the different end-goals (the height of the blocks). 
During real-time execution, 
we included a dynamic obstacle in front of the third, and last, block. 
The demonstrated path and the executed path for this condition are shown in Fig.~\ref{fig:task_stacking_paths}. 

\begin{figure}[t]
\centering
    {\scriptsize
\def\svgwidth{0.95\columnwidth} 
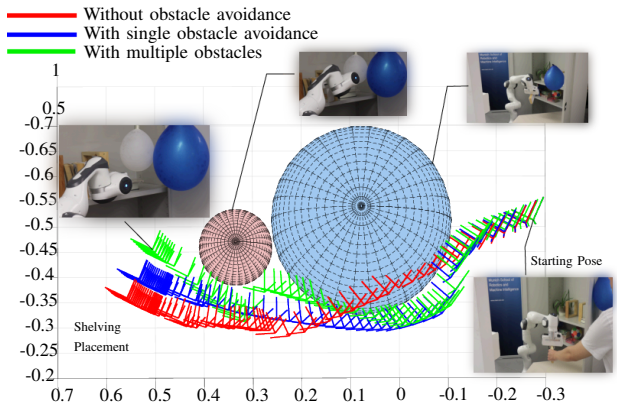
} 
\caption{Top view of the shelving task paths showing in Fig.~\ref{fig:tasks_front_page}. Deflection of the final computed path (blue, green) is crucial for this complex scenario. Note that the plan without collision avoidance (red) is inside the blue obstacle.}
\label{fig:shelf_exp}
\end{figure}%

\subsection{Reactive user-guided motion planning in clutter}

To highlight the performance and robustness of our approach,  
we devised a cluttered scene where the objective was to shelve books handed by a human 
in a cluttered cupboard with a specific attitude transformation.  
In addition to static obstacles, 
the robot also avoids two unforeseen obstacles 
that appear during the placing of the second and the third book.  
Notice the demonstration was provided in a different shelf (lower shelf) as shown in Fig.~\ref{fig:shelf_strip}. 
Regardless, 
the proposed framework enabled real-time collision-free solution directly with one single new information: the new goal pose. 
The resulting trajectory for the three book placements are shown in Fig.~\ref{fig:shelf_exp}.   
The red, blue and green curves depict the paths to place three books in a line. The first path is without any obstacle, the second one is with one obstacle (the larger one) and the third is with two obstacles of different size.

\begin{table}[t]
\footnotesize
\caption{The average computation time for a complete path}
\begin{tabular}{cccc}
 & Pouring & Stacking & Shelving\tabularnewline
\cline{2-4} \cline{3-4} \cline{4-4} 
w/o REPET. & $16.464\pm2.909$ & $6.448\pm0.934$ & $5.185\pm1.313$\tabularnewline
REPET & $16.364\pm3.004$ & $4.287\pm0.800$ & $5.792\pm0.644$\tabularnewline
\hline 
\end{tabular}
\label{table:computationTime}
\end{table}

\subsection{Analyze the real-time capabilities}

The average computation time to complete all the aforementioned tasks \footnote{The average time refers to the complete task and not to the motion-generation and control loop which were running under $0.1$ ms for all cases.} are shown in Table~\ref{table:computationTime}. 
%
%
%
Notice that for some tasks, the collision avoidance led to less computational time. This was due to the deflected trajectory being closer to the goal and highlights the heuristic designed in Section \ref{sec:obstacle_avoidance}. 
%
This also highlights the real-time capability of our approach. 
The computation for each rapidly expanding plane-oriented escaping tree level   took approximately $0.0429\pm0.01094$ ms, therefore having the potential to evade dynamic obstacles in real-time.
Algorithms were implemented in C++ (not optimized) using the DQ\_Robotics \cite{Adorno2019_dqrobotics} library. 

\subsection{Transfer to different kinematics: Bi-manual System}
\begin{figure}[t]
\centering
{\scriptsize
\def\svgwidth{0.98\columnwidth} 
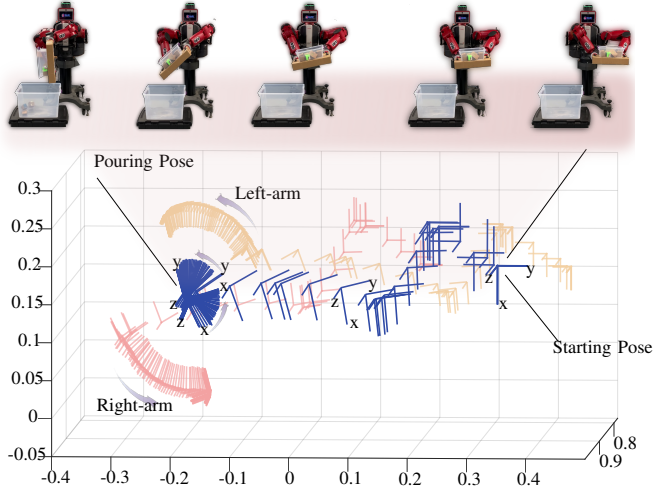
}    
\caption{Transfer generalization between user-guided demonstration from single-arm to the Baxter dual-arm robot for pouring/transfer task. The screenshots on the top depicts different moments of the task from right to left. The yellow and red poses depict the path for the left and right end-effectors whereas planned path using the single-arm guidance is shown in blue.  Notice that both end-effectors need to combine coupled attitude and translation motion to ensure the final pouring path consists of mostly rotation as shown in the single-arm demonstration.}
\label{fig:transfergrasps}
\end{figure}

The final experiment that we conduct validates one interesting feature of our work -- generalizing to multiple robots with different hardware architecture and kinematic chain. As our user guidance based planning is done in $\unitdualquatgroup$, it implies that the same user demonstration can be used again for roughly same constrained tasks and/or for different robotic systems. Therefore, we used the demonstration provided on the Panda robot for the pouring task and used the final generated path to a new goal location as an input for the dual-arm Baxter transfer task   (Fig.~\ref{fig:transfergrasps}) which also inherently had different initial and final poses. 

\section{Conclusion} 
\label{sec:conclusion}

This paper presents 
a novel framework for using human demonstration to generate real-time motion plans for complex tasks that satisfy the embedded implicit constraints. Our approach includes a reactive collision-avoidance algorithm that allows execution in the presence of obstacles that were not present during demonstration. Our technical approach exploits the group structure of rigid body motion and also utilizes the computational efficiency of $\unitdualquatgroup$ representation of rigid body configuration. This allows real-time execution. Extensive experimental studies were performed to highlight the generalization capability of our approach. In future works, we plan to integrate vision feedback for unknown environments \cite{2021_Farias_Maxime_Brahim_Stolkin_Naresh__CASE}, improve smoothness to C$1$ trajectories \cite{Allmendinger2018}, and introduce multi-modal strategies and human-factors for safe online trajectory adaptation in HRI scenarios as in \cite{2018_Lipeng_Humanoids}.


\bibliographystyle{IEEEtran}
\bibliography{myreferences.bib,bibtexCoopAndOthers.bib}

\begin{thebibliography}{10}
\providecommand{\url}[1]{#1}
\csname url@samestyle\endcsname
\providecommand{\newblock}{\relax}
\providecommand{\bibinfo}[2]{#2}
\providecommand{\BIBentrySTDinterwordspacing}{\spaceskip=0pt\relax}
\providecommand{\BIBentryALTinterwordstretchfactor}{4}
\providecommand{\BIBentryALTinterwordspacing}{\spaceskip=\fontdimen2\font plus
\BIBentryALTinterwordstretchfactor\fontdimen3\font minus
  \fontdimen4\font\relax}
\providecommand{\BIBforeignlanguage}[2]{{%
\expandafter\ifx\csname l@#1\endcsname\relax
\typeout{** WARNING: IEEEtran.bst: No hyphenation pattern has been}%
\typeout{** loaded for the language `#1'. Using the pattern for}%
\typeout{** the default language instead.}%
\else
\language=\csname l@#1\endcsname
\fi
#2}}
\providecommand{\BIBdecl}{\relax}
\BIBdecl

\bibitem{kragic}
D.~Kragic, J.~Gustafson, H.~Karaoguz, P.~Jensfelt, and R.~Krug, ``Interactive,
  collaborative robots: Challenges and opportunities.'' in \emph{IJCAI}, 2018.

\bibitem{laha21cooperative}
R.~Laha, L.~F. Figueredo, J.~Vrabel, A.~Swikir, and S.~Haddadin, ``{Reactive
  Cooperative Manipulation based on Set Primitives and Circular Fields},'' in
  \emph{IEEE International Conference on Robotics and Automation}, Xi'an,
  China, May 2021.

\bibitem{Laha_Thesis}
R.~Laha, ``Task-specific motion planning using user-guidance, imitation, and
  self evaluation,'' Master's thesis, Stony Brook University, New York, 2018.

\bibitem{ravichandar2020recent}
H.~Ravichandar, A.~S. Polydoros, S.~Chernova, and A.~Billard, ``Recent advances
  in robot learning from demonstration,'' \emph{Annual Review of Control,
  Robotics, and Autonomous Systems}, vol.~3, pp. 297--330, 2020.

\bibitem{calinon2018learning}
S.~Calinon, ``Learning from demonstration (programming by demonstration),''
  \emph{Encyclopedia of robotics}, pp. 1--8, 2018.

\bibitem{ijspeert2013dynamical}
A.~J. Ijspeert, J.~Nakanishi, H.~Hoffmann, P.~Pastor, and S.~Schaal,
  ``Dynamical movement primitives: learning attractor models for motor
  behaviors,'' \emph{Neural computation}, vol.~25, no.~2, pp. 328--373, 2013.

\bibitem{Ijspeert2002}
A.~Ijspeert, J.~Nakanishi, and S.~Schaal, ``Movement imitation with nonlinear
  dynamical systems in humanoid robots,'' in \emph{IEEE International
  Conference on Robotics and Automation (ICRA)}, 2002.

\bibitem{paraschos2013promp}
A.~Paraschos, C.~Daniel, J.~R. Peters, and G.~Neumann, ``Probabilistic movement
  primitives,'' in \emph{Advances in Neural Information Processing Systems},
  C.~J.~C. Burges, L.~Bottou, M.~Welling, Z.~Ghahramani, and K.~Q. Weinberger,
  Eds., vol.~26.\hskip 1em plus 0.5em minus 0.4em\relax Curran Associates,
  Inc., 2013.

\bibitem{Silverio2017}
J.~Silv{\'{e}}rio, S.~Calinon, L.~Rozo, and D.~G. Caldwell, ``{Learning task
  priorities from demonstrations},'' \emph{IEEE Transactions on Robotics},
  vol.~35, no.~1, pp. 78--94, 2019.

\bibitem{silverio2017learning}
J.~Silv{\'e}rio, S.~Calinon, L.~Rozo, and D.~G. Caldwell, ``Learning competing
  constraints and task priorities from demonstrations of bimanual skills,''
  \emph{arXiv preprint}, 2017.

\bibitem{WuD10}
Y.~Wu and Y.~Demiris, ``Towards one shot learning by imitation for humanoid
  robots,'' in \emph{ICRA}.\hskip 1em plus 0.5em minus 0.4em\relax IEEE, 2010.

\bibitem{Zeestraten17RAL}
M.~Zeestraten, I.~Havoutis, J.~Silv\'erio, S.~Calinon, and D.~Caldwell, ``An
  approach for imitation learning on {R}iemannian manifolds,'' \emph{{RA-L}},
  2017.

\bibitem{YuFX+18}
T.~Yu, C.~Finn, A.~Xie, S.~Dasari, T.~Zhang, P.~Abbeel, and S.~Levine,
  ``One-shot imitation from observing humans via domain-adaptive
  meta-learning,'' in \emph{Robotics:Science \& Systems}, 2018.

\bibitem{atkeson_schaal_1997}
C.~Atkeson and S.~Schaal, ``Learning tasks from a single demonstration,'' in
  \emph{Proceedings of International Conference on Robotics and Automation},
  vol.~2, 1997, pp. 1706--1712 vol.2.

\bibitem{praveena2019user}
P.~Praveena, D.~Rakita, B.~Mutlu, and M.~Gleicher, ``User-guided offline
  synthesis of robot arm motion from 6-dof paths,'' in \emph{2019 International
  Conference on Robotics and Automation (ICRA)}.\hskip 1em plus 0.5em minus
  0.4em\relax IEEE, 2019, pp. 8825--8831.

\bibitem{denny2016theory}
J.~Denny, J.~Colbert, H.~Qin, and N.~M. Amato, ``On the theory of user-guided
  planning,'' in \emph{2016 IEEE/RSJ International Conference on Intelligent
  Robots and Systems (IROS)}.\hskip 1em plus 0.5em minus 0.4em\relax IEEE,
  2016, pp. 4794--4801.

\bibitem{phillips1990interactive}
C.~B. Phillips, J.~Zhao, and N.~I. Badler, ``Interactive real-time articulated
  figure manipulation using multiple kinematic constraints,'' in
  \emph{Proceedings of the 1990 Symposium on interactive 3D Graphics}, 1990,
  pp. 245--250.

\bibitem{gleicher1998retargetting}
M.~Gleicher, ``Retargetting motion to new characters,'' in \emph{Proceedings of
  the 25th annual conference on Computer graphics and interactive techniques},
  1998, pp. 33--42.

\bibitem{islam2018online}
F.~Islam, O.~Salzman, and M.~Likhachev, ``Online, interactive user guidance for
  high-dimensional, constrained motion planning,'' in \emph{Proceedings of the
  27th International Joint Conference on Artificial Intelligence}, 2018, pp.
  4921--4928.

\bibitem{pham2013kinodynamic}
Q.-C. Pham, S.~Caron, and Y.~Nakamura, ``Kinodynamic planning in the
  configuration space via admissible velocity propagation.'' in \emph{Robotics:
  Science and Systems}, vol.~32, 2013.

\bibitem{nakanishi2003learning}
J.~Nakanishi, J.~Morimoto, G.~Endo, G.~Cheng, S.~Schaal, and M.~Kawato,
  ``Learning from demonstration and adaptation of biped locomotion with
  dynamical movement primitives,'' in \emph{Workshop on Robot Programming by
  Demonstration, IEEE/RSJ International Conference on Intelligent Robots and
  Systems}, 2003.

\bibitem{Adorno2017}
B.~V. Adorno, ``{Robot Kinematic Modeling and Control Based on Dual Quaternion
  Algebra -- Part I: Fundamentals} - hal-01478225,'' p.~47, 2017.

\bibitem{2011_Adorno_THESIS}
B.~V. A, ``Two-arm manipulation: From manipulators to enhanced human-robot
  collaboration,'' Ph.D. dissertation, Laboratoire d'Informatique, de Robotique
  et de Micro\'{e}lectronique de Montpellier (LIRMM) - Universit\'{e}
  Montpellier 2, Montpellier, France, 2011.

\bibitem{2005_Wu_Hu_Xu_Li_Lian_TAES}
Y.~Wu, X.~Hu, D.~Hu, T.~Li, and J.~Lian, ``{Strapdown inertial navigation
  system algorithms based on dual quaternions},'' \emph{IEEE Transactions On
  Aerospace And Electronic Systems}, vol.~41, no.~1, pp. 110--132, 2005.

\bibitem{2016_Figueredo_PhDThesis}
L.~F.~C. Figueredo, ``Kinematic control based on dual quaternion algebra and
  its application to robot manipulators,'' Ph.D. dissertation, University of
  Brasilia, Brazil, 2016.

\bibitem{kussaba2017hybrid}
H.~T. Kussaba, L.~F. Figueredo, J.~Y. Ishihara, and B.~V. Adorno, ``Hybrid
  kinematic control for rigid body pose stabilization using dual quaternions,''
  \emph{Journal of the Franklin Institute}, vol. 354, no.~7, pp. 2769--2787,
  2017.

\bibitem{magro2017dual}
P.~P. Magro, H.~T. Kussaba, L.~F. Figueredo, and J.~Y. Ishihara, ``Dual
  quaternion-based bimodal global control for robust rigid body pose kinematic
  stabilization,'' in \emph{American Control Conference (ACC), 2017}.\hskip 1em
  plus 0.5em minus 0.4em\relax IEEE, 2017, pp. 1205--1210.

\bibitem{1990_Funda_Paul_TRA}
J.~Funda and R.~Paul, ``{A computational analysis of screw transformations in
  robotics},'' \emph{IEEE Transactions on Robotics and Automation}, vol.~6,
  no.~3, pp. 348--356, 1990.

\bibitem{OZGUR201666}
\BIBentryALTinterwordspacing
E.~Özgür and Y.~Mezouar, ``Kinematic modeling and control of a robot arm
  using unit dual quaternions,'' \emph{Robotics and Autonomous Systems},
  vol.~77, pp. 66 -- 73, 2016. [Online]. Available:
  \url{http://www.sciencedirect.com/science/article/pii/S0921889015301184}
\BIBentrySTDinterwordspacing

\bibitem{2019_Xialong_etal__JIRS__DQefficiency_invDyn_Parallel}
X.~Yang, H.~Wu, Y.~Li, S.~Kang, and B.~Chen, ``Computationally efficient
  inverse dynamics of a class of six-dof parallel robots: Dual quaternion
  approach,'' \emph{Journal of Intelligent \& Robotic Systems}, vol.~94, pp.
  101--113, 2019.

\bibitem{1998_Aspragathos_Dimitros_TSMC}
N.~A. Aspragathos and J.~K. Dimitros, ``{A Comparative Study of Three Methods
  for Robot Kinematics},'' \emph{IEEE Transactions on Systems, Man, and
  Cybernetics, Part B: Cybernetics}, vol.~28, no.~2, pp. 135--145, 1998.

\bibitem{kuipers:1999}
J.~Kuipers, \emph{{Quaternions and Rotation Sequences: A Primer with
  Applications to Orbits, Aerospace, and Virtual Reality}}.\hskip 1em plus
  0.5em minus 0.4em\relax Princeton University Press, 1999.

\bibitem{Selig2005}
J.~M. Selig, \emph{Geometric Fundamentals of Robotics}, 2nd~ed.\hskip 1em plus
  0.5em minus 0.4em\relax Springer-Verlag New York Inc., 2005.

\bibitem{Book:Boothby:2002}
W.~M. Boothby, \emph{An Introduction to Differentiable Manifolds and Riemannian
  Geometry}, 2nd~ed.\hskip 1em plus 0.5em minus 0.4em\relax Academic Press,
  2002.

\bibitem{2017_Busam_Birdal_Navab_ICCVW}
B.~Busam, T.~Birdal, and N.~Navab, ``Camera pose filtering with local
  regression geodesics on the riemannian manifold of dual quaternions,'' in
  \emph{2017 IEEE International Conference on Computer Vision (ICCV)}, 2017,
  pp. 2436--2445.

\bibitem{1995_Park_JMD_ASME}
F.~C. Park, ``{Distance Metrics on the Rigid-Body Motions with Applications to
  Mechanism Design},'' \emph{Journal of Mechanical Design -- Transactions of
  ASME}, vol. 117, no.~1, pp. 48--54, 1995.

\bibitem{Zacur2014b}
E.~Zacur, M.~Bossa, and S.~Olmos, ``{Left-Invariant Riemannian Geodesics on
  Spatial Transformation Groups},'' \emph{SIAM Journal on Imaging Sciences},
  vol.~7, no.~3, pp. 1503--1557, jul 2014.

\bibitem{2009_Sachkov_JMS}
Y.~L. Sachkov, ``Control theory on {L}ie groups,'' \emph{Journal of
  Mathematical Sciences}, vol. 156, no.~3, pp. 381--439, 2009.

\bibitem{BOOK:2016:Gallier_Quaintance}
J.~Gallier and J.~Quaintance, \emph{Notes on Differential Geometry and Lie
  Groups}.\hskip 1em plus 0.5em minus 0.4em\relax Department of Computer and
  Information Science University of Pennsylvania, 2016.

\bibitem{2017_Busam_Birdal_Navab__ArXiv}
\BIBentryALTinterwordspacing
B.~Busam, T.~Birdal, and N.~Navab, ``{Camera Pose Filtering with Local
  Regression Geodesics on the Riemannian Manifold of Dual Quaternions},''
  \emph{ArXiv e-prints}, 2017. [Online]. Available:
  \url{http://arxiv.org/abs/1704.07072}
\BIBentrySTDinterwordspacing

\bibitem{2013_Lorenzi_Pennec__IJCV}
M.~Lorenzi and X.~Pennec, ``Geodesics, parallel transport \& one-parameter
  subgroups for diffeomorphic image registration,'' \emph{International Journal
  of Computer Vision}, vol. 105, pp. 111--127, 2013.

\bibitem{sarker2020screw}
A.~Sarker, A.~Sinha, and N.~Chakraborty, ``On screw linear interpolation for
  point-to-point path planning,'' in \emph{2020 IEEE/RSJ International
  Conference on Intelligent Robots and Systems (IROS)}.\hskip 1em plus 0.5em
  minus 0.4em\relax IEEE, 2020, pp. 9480--9487.

\bibitem{2012_Wang_Han_Yu_Zheng_JMAA}
X.~Wang, D.~Han, C.~Yu, and Z.~Zheng, ``{The geometric structure of unit dual
  quaternion with application in kinematic control},'' \emph{Journal of
  Mathematical Analysis and Applications}, vol. 389, no.~2, pp. 1352--1364, May
  2012.

\bibitem{DQBlending}
L.~Kavan, S.~Collins, C.~O'Sullivan, and J.~Zara, ``Dual quaternions for rigid
  transformation blending,'' Trinity College Dublin, Tech. Rep., Trinity
  College Dublin 2006.

\bibitem{article:1996_Kim_Kim_Shin__JVCA}
M.-j. Kim, M.-s. Kim, and S.~Y. Shin, ``A compact differential formula for the
  first derivative of a unit quaternion curve,'' \emph{The Journal of
  Visualization and Computer Animation}, vol.~7, no.~1, pp. 43--57, 1996.

\bibitem{Allmendinger2018}
F.~Allmendinger, S.~{Charaf Eddine}, and B.~Corves, ``{Coordinate-invariant
  rigid-body interpolation on a parametric C1 dual quaternion curve},''
  \emph{Mechanism and Machine Theory}, pp. 731--744, March 2018.

\bibitem{grassmann2018smooth}
R.~Grassmann, L.~Johannsmeier, and S.~Haddadin, ``Smooth point-to-point
  trajectory planning in $se(3)$ with self-collision and joint constraints
  avoidance,'' in \emph{2018 IEEE/RSJ International Conference on Intelligent
  Robots and Systems (IROS)}.\hskip 1em plus 0.5em minus 0.4em\relax IEEE,
  2018, pp. 1--9.

\bibitem{laha21p2p}
R.~Laha, A.~Rao, L.~Figueredo, Q.~Chang, S.~Haddadin, and N.~Chakraborty,
  ``Point-to-point path planning based on user guidance and screw linear
  interpolation,'' in \emph{Proceedings of the ASME International Design
  Engineering Technical Conferences and Computers and Information in
  Engineering Conference (IDETC/CIE)}, August 2021.

\bibitem{agrawal1987hamilton}
O.~P. Agrawal, ``Hamilton operators and dual-number-quaternions in spatial
  kinematics,'' \emph{Mechanism and machine theory}, vol.~22, no.~6, pp.
  569--575, 1987.

\bibitem{2013_Figueredo_Adorno_Ishihara_Borges_ICRA}
L.~Figueredo, B.~Adorno, J.~Ishihara, and G.~Borges, ``Robust kinematic control
  of manipulator robots using dual quaternion representation,'' in \emph{IEEE
  International Conference on Robotics and Automation (ICRA)}, 2013, pp.
  1949--1955.

\bibitem{2001_Han_Park_TAC}
Y.~Han and F.~Park, ``{Least squares tracking on the Euclidean group},''
  \emph{IEEE Transactions on Automatic Control}, vol.~46, no.~7, pp.
  1127--1132, 2001.

\bibitem{Adorno2019_dqrobotics}
B.~V. {Adorno} and M.~M. {Marinho}, ``Dq robotics: A library for robot modeling
  and control,'' \emph{IEEE Robotics Automation Magazine}, 2020.

\bibitem{2021_Farias_Maxime_Brahim_Stolkin_Naresh__CASE}
C.~De~Farias, M.~Adjigble, B.~Tamadazte, R.~Stolkin, and N.~Marturi, ``Dual
  quaternion-based visual servoing for grasping moving objects,'' in \emph{2021
  IEEE 17th International Conference on Automation Science and Engineering
  (CASE)}, 2021, pp. 151--158.

\bibitem{2018_Lipeng_Humanoids}
L.~{Chen}, L.~F. {Figueredo}, and M.~R. {Dogar}, ``Planning for muscular and
  peripersonal-space comfort during human-robot forceful collaboration,'' in
  \emph{2018 IEEE-RAS 18th International Conference on Humanoid Robots
  (Humanoids)}, Nov 2018, pp. 1--8.

\end{thebibliography}
\end{document}